\pdfoutput=1

\documentclass[11pt]{article}

\PassOptionsToPackage{table}{xcolor}
\usepackage[final]{acl}

\usepackage{times}
\usepackage{booktabs}
\usepackage{multirow}
\usepackage{booktabs}
\usepackage{caption}
\usepackage{latexsym}
\usepackage{floatrow}
\usepackage{amsmath}  
\usepackage{amsfonts}
\usepackage{amssymb}
\usepackage{pifont}

\usepackage{float}

\usepackage{listings}

\usepackage[T1]{fontenc}

\usepackage[utf8]{inputenc}

\usepackage{microtype}

\usepackage{inconsolata}

\usepackage{graphicx}
\usepackage{makecell}

\newcommand{\cmark}{\ding{51}}  
\newcommand{\xmark}{\ding{55}}  
\title{SPECS: Specificity-Enhanced CLIPScore \\ for  Long Image Caption Evaluation}

\author{
  Xiaofu Chen$^{1}$ \quad
  Israfel Salazar$^{2}$ \quad
  Yova Kementchedjhieva$^{1}$ \\
  $^{1}$MBZUAI \quad
  $^{2}$University of Copenhagen \\
  \texttt{\{xiaofu.chen, yova.kementchedjhieva\}@mbzuai.ac.ae} \\
  \texttt{israfel.salazar@di.ku.dk}
}

\begin{document}
\maketitle

\begin{abstract}
As interest grows in generating long, detailed image captions, standard  evaluation metrics become increasingly unreliable. N-gram-based metrics though efficient, fail to capture semantic correctness. Representational Similarity (RS) metrics, designed to address this, initially saw limited use due to high computational costs, while today, despite advances in hardware, they remain unpopular due to low correlation to human judgments. Meanwhile, metrics based on large language models (LLMs) show strong correlation with human judgments, but remain too expensive for iterative use during model development.
We introduce SPECS (Specificity-Enhanced CLIPScore), a reference-free RS metric tailored to long image captioning. SPECS modifies CLIP with a new objective that emphasizes specificity: rewarding correct details and penalizing incorrect ones. We show that SPECS matches the performance of open-source LLM-based metrics in correlation to human judgments, while being far more efficient. This makes it a practical alternative for iterative checkpoint evaluation during image captioning model development.
Our code can be found at \url{https://github.com/mbzuai-nlp/SPECS}.

\end{abstract}

\section{Introduction}
Image captioning has been a key topic in vision-language research, offering a controlled setting to study grounded language generation~\cite{karpathy2015deep}. While early efforts focused on short, general captions~\cite{vinyals2015show}, recent work has shifted toward generating long, detailed descriptions that capture fine-grained visual information\cite{johnson2016densecap, cho-etal-2022-fine,doveh2023dense,li2023blip}. This complex task requires strong visual grounding and improved cross-modal alignment \cite{ liu2024payingattentionimagetrainingfree, li2021align, xie2025fg}. It expands the scope of generative vision-language modeling but also magnifies a long-standing challenge: how to evaluate captions reliably and efficiently.

\begin{figure}
    \centering
    \includegraphics[width=1\linewidth]{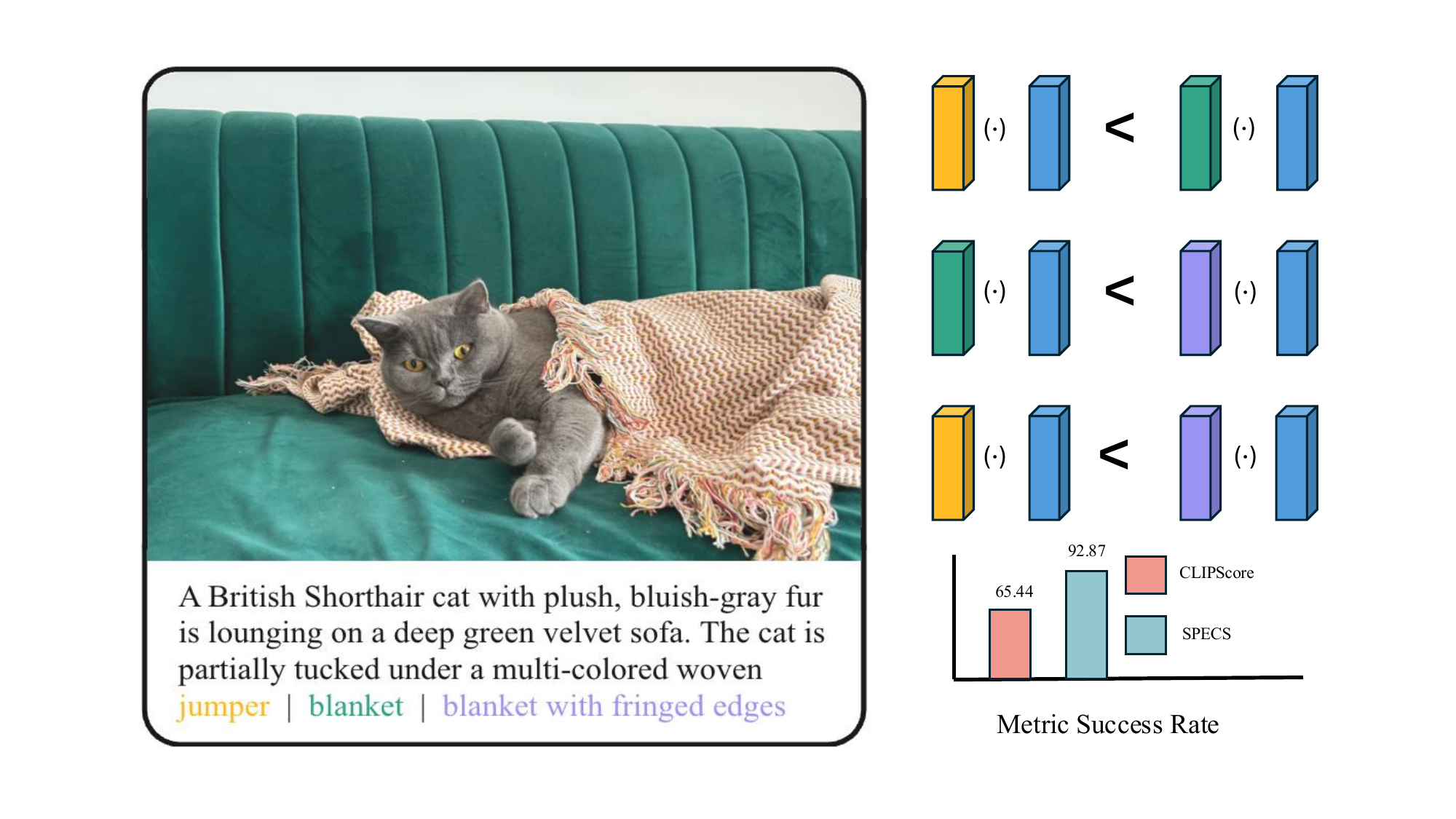}
    \caption{Example of specificity in caption evaluation. Given an image, captions with increasing correct details (“jumper” $\rightarrow$ “blanket” $\rightarrow$ “blanket with fringed edges”) should receive progressively higher cosine similarity, $(\cdot)$, to the image. SPECS ranks these minimal pairs correctly at a high rate, reflecting its strong specificity.
    }
    \label{fig:first_intuition}
\end{figure} 

Automatic evaluation in captioning, as in other natural language generation tasks, has long been a challenge~\cite{otani2023toward, wang2023delving}. Early metrics rely on $n$-gram overlap between generated captions and references. While computationally efficient, these methods fail to capture semantic similarity, often penalizing valid paraphrases and underestimating the severity of hallucinations, even in short captions \cite{papineni2002bleu,banerjee2005meteor,vedantam2015cider,lin2004rouge}. 

To address these limitations, Representational Similarity (RS) metrics use pretrained vision-language models to compare image and caption embeddings in a shared feature space~\cite{hessel2021clipscore,sarto2023positive}, often eliminating the need for reference captions. More recently, evaluation metrics based on large language models (LLMs) have become the standard, showing strong correlation with human judgments, especially as the length of generated image captions increases \cite{chan2023clair,yu2024rlaif,ye2025painting}.

Each new generation of evaluation metrics brings improvements in semantic expressiveness, but often at the cost of higher computational requirements. As a result, there is frequently a mismatch between what is technically feasible and what is practical for routine model development. For instance, although CLIPScore~\cite{hessel2021clipscore} demonstrated stronger semantic alignment than traditional metrics like CIDEr~\cite{vedantam2015cider}, its adoption remained limited due to the high computational cost at the time of its release. A similar pattern is now emerging with LLM-based metrics: while they achieve state-of-the-art correlation with human judgments, their inference cost makes them impractical for iterative evaluation during model training and development.

This tradeoff between reliability and efficiency becomes particularly problematic in the context of long image captioning, where no existing metric can strike a reasonable balance between the two. Recent studies  have shown that even simple metrics like BLEU-4 outperform RS metrics like CLIPScore in terms of correlation with human judgments \cite[sample-level Kendall's Tau correlations of 0.27 and 0.17, respectively]{ye2025painting}. These results underscore a pressing gap: there is no metric that reliably evaluates the quality of long captions while remaining computationally practical.

In this work, we introduce \textbf{SPECS} (\textbf{SP}ecificity-\textbf{E}nhanced \textbf{C}LIP-\textbf{S}core), a reference-free RS metric designed for the evaluating long image captions. SPECS builds on a long-context adaptation of CLIP ~\cite{zhang2024long} and incorporates a new training objective that emphasizes \textit{specificity}: rewarding correct details and penalizing incorrect ones. In extensive evaluations, SPECS matches the best open-source LLM-based metric~\cite{lee2024fleur} in terms of correlation to human judgments, while being over two orders of magnitude more efficient. SPECS is a practical and scalable solution for iterative model development in long caption generation.

\section{Related work}
\label{sec:rw}

Image captioning metrics are central to evaluating vision language models (VLMs). These metrics score how well a model can describe an image, in a way that aligns with human judgment. 

\paragraph{Image Caption Evaluation Methods}  
Early work relied on $n$-gram matching, where metrics such as BLEU~\cite{papineni2002bleu}, METEOR~\cite{banerjee2005meteor}, ROUGE~\cite{lin2004rouge}, and CIDEr~\cite{vedantam2015cider} compared generated captions against human references using surface lexical overlap. While easy to compute, these metrics often fail when captions use different but valid phrasings, leading to low correlation with human judgment. To overcome this, later approaches explored semantic parsing, such as SPICE~\cite{anderson2016spice}, which evaluates scene-graph structures. CLIPScore~\cite{hessel2021clipscore} and PACScore~\cite{sarto2023positive} build on pre-trained vision–language models such as CLIP~\cite{radford2021learning}, assessing captions through representational similarity between images and their captions. 

More recently, large language models (LLMs) have been used as evaluators by prompting them to score alignment between captions and images: FaithScore~\cite{jing2311faithscore} extracts atomic facts from captions using an MLLM and verifying each fact against the image to measure faithfulness. CLAIR~\cite{chan2023clair} leverages LLMs in a zero-shot setting to score captions and explain their judgments,  GPT4-Eval~\cite{liu2023visual} is evaluated by prompting GPT-4 to judge multimodal responses, with performance measured by relative quality scores. RLAIF-V~\cite{yu2024rlaif} evaluates captions by prompting open-source MLLMs to verify decomposed claims and score hallucination reduction. FLEUR~\cite{lee2024fleur} prompts an open-source MLLM to directly compare an image with a candidate caption in a reference-free setting, producing both a numerical score and a natural-language rationale to align evaluation with human judgments. CAPTURE~\cite{dong2024benchmarking} evaluates captions by extracting and matching fine-grained visual details. DCScore~\cite{ye2025painting} evaluates captions by decomposing them into primitive information units and measuring their accuracy and coverage using GPT-4. These LLM-based methods show strong performance but remain prohibitively expensive and mostly closed-source.


\paragraph{CLIP-based adaptations}
Metrics based on representational similarity carry promise as they operate on the level of semantics rather than surface, and they are less costly than LLMs.
However, CLIP, the common choice for base model in such metrics, is not suitable for long caption evaluation for two key reasons: it lacks compositionality and it can only take up to 77 subword tokens.  To address these limitations, a line of research has modified CLIP in various ways. NegCLIP~\cite{yuksekgonul2022and} is the first work to address compositionality, enhancing robustness by penalizing captions that introduce wrong attributes and reducing the risk of rewarding hallucinated details. Later, LaCLIP~\cite{fan2023improving} and TripletCLIP~\cite{patel2024tripletclip} further advance compositional evaluation, making metrics more sensitive to how attributes and objects are combined. DAC~\cite{doveh2023dense} fine-tunes CLIP on enhanced captions to improve caption density and strengthen compositional reasoning. DCI~\cite{urbanek2024picture} introduces a densely captioned dataset with long, region-aligned descriptions and trains CLIP for improved long-text understanding. 
To address the issue of context length, LongCLIP~\cite{zhang2024long} modifies CLIP to process longer textual inputs through interpolation of  the positional embeddings. LongCLIP exhibits strong long-caption understanding in retrieval tasks. The models listed above were not designed as metrics as such, but we hypothesize that their intended strengths might benefit long caption evaluation.

Recent evaluation metrics have made progress, but they still face two main problems. Some metrics do not capture detailed information well, while others do so at a high computational cost. We propose a new metric which leverages representational similarity, with a focus on specificity---the ability of a vision-language model to consistently prefer more informative, visually grounded captions at varying caption lengths.

\section{Specificity}
Let us consider the three caption variants depicted in Figure~\ref{fig:first_intuition}. Describing the cat as tucked under a \textit{blanket} is correct, and because the caption already contains other relevant details, a good evaluation metric should assign it a high score. If the caption further specified that the blanket has \textit{fringed edges}, the score should increase slightly, reflecting the  correct additional detail. On the other hand, if instead of a \textit{blanket} the caption said that the cat was lying under a \textit{jumper}, that should result in a slightly lower score---most of the details remain accurate, but this particular object mentioned is incorrect. This simple example illustrates the notion of \textit{specificity} which we adapt from \citet{xu2024benchmarking} to mean: the ability of a text representation to encode every detail in a caption in a way that correctly reflects the relevance of this detail to a reference image.
A metric based on a specificity-enhanced model would thus favor captions that include more relevant details and penalize those that omit important information or introduce hallucinated or erroneous content. Such a metric implicitly  implements the notions of soft precision and recall.

\subsection{Detail Units}\label{subsec:detailunit}
To concretely evaluate specificity, we begin by introducing the key concept of a \textbf{detail unit}. The abstract notion behind a detail unit refers to any minimal bit of information in a caption, such as the presence of a \textit{blanket}, the \textit{fringed edges} of the blanket, etc. For operational purposes, however, we define a detail unit to mean a phrase which contributes at least one new visual detail (and possibly more), and fits syntactically and semantically within the preceding context. Under this definition, \textit{a blanket} is a detail unit, and so is \textit{a blanket with fringed edges},  but \textit{a blanket with} is not, and neither is \textit{The cat} in the middle of the caption in Figure~\ref{fig:first_intuition}, since it does not contribute new information.

Formally, we denote an image-caption pair as $\{i,c\}$, and decompose a caption as $c = \{d_1, d_2, \dots, d_m\}$ where each $d_i$ is a detail unit.
Every subsequence of detail units, built cumulatively from left to right, constitutes a valid caption: {$c_1 = \{d_1\}$, $c_2 = \{d_1 + d_2\}$, ..., $\{c = d_1 + \dots + d_m \}$, each containing progressively more information. 
Given a ground-truth, high-quality caption, this ordered sequence should exhibit monotonically increasing representational similarity to its reference image, under a specificity-enhanced model. Conversely, if an incorrect detail unit is added at any point, this should be reflected in a dip in the similarity score. 
This decomposition provides a structured way to test and enhance model specificity to visual detail across any caption length.

\begin{figure*}[t]
    \centering
    \includegraphics[width=1.\linewidth]{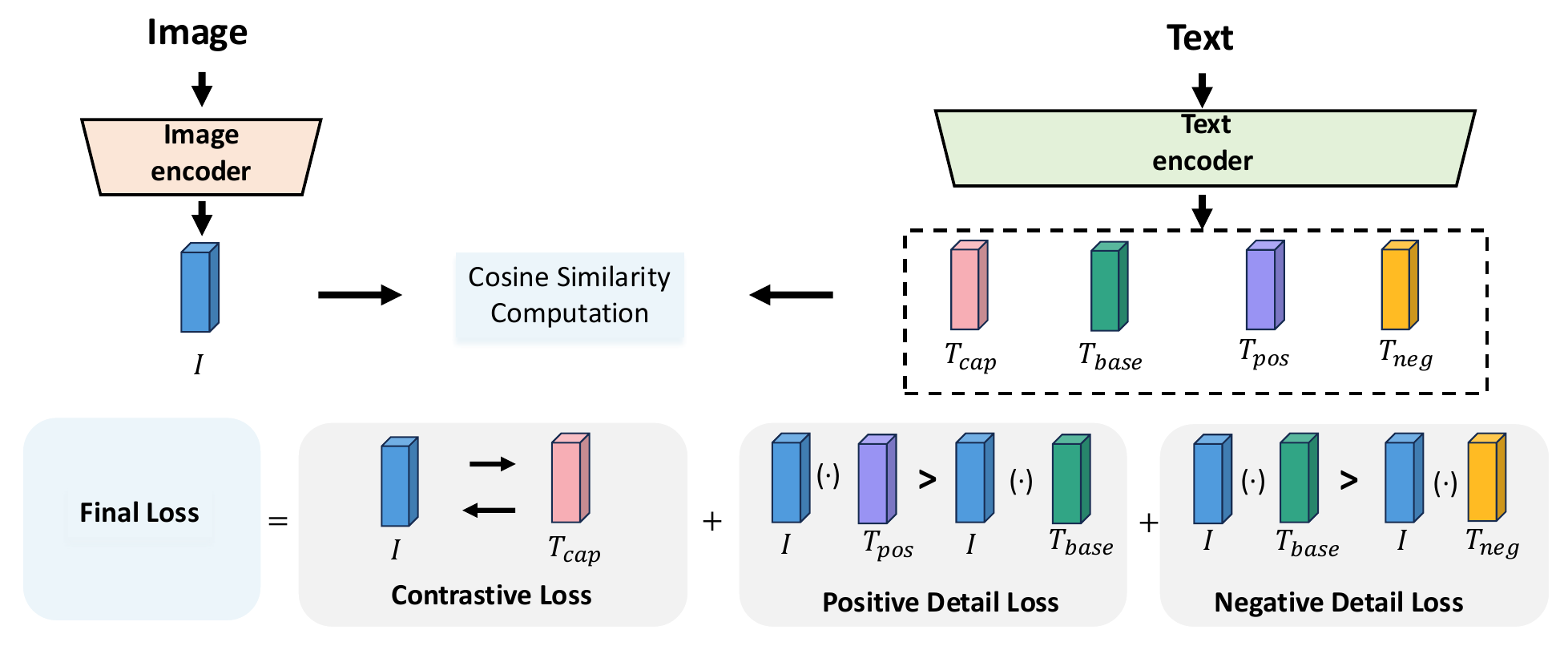}
    \caption{Training framework. Given an image and its caption, we produce a base caption, a more positive caption, and a negative caption. The model computes embeddings and is trained with three losses: a contrastive loss $\mathcal{L}_{\text{contrastive}}$ on the full caption, a positive detail loss $\mathcal{L}_{\text{pos}}$ to prefer more informative descriptions, and a negative detail loss $\mathcal{L}_{\text{neg}}$ to penalize misleading ones. This setup encourages sensitivity to fine-grained textual differences. The symbol $(\cdot)$ denotes cosine similarity computation. Here, $I$ is the image embedding, $T_{\text{cap}}$ is the original caption, $T_{\text{base}}$ is the base caption, $T_{\text{pos}}$ is the more detailed caption, and $T_{\text{neg}}$ is the negative caption.}
    \label{fig:model-fig}
\end{figure*}

Detail units that contain relevant information are referred to as \textbf{positive} ($d_+$), while detail units that introduce content not grounded in the image are referred to as \textbf{negative} ($d_-$). The expected behavior of a model with good specificity is then to assign higher similarity to the pair $\{i,c_j + d_+\}$ than to the pair $\{i,c_j\}$, and a lower similarity to the pair $\{i,c_j + d_-\}$ than to the pair $\{i,c_j\}$, where $j \in [1,...,m]$ and $c_j$ is a partial caption for the image. Each triplet, $\{i, c_j, c_j + d_+\}$ and $\{i, c_j, c_j + d_-\}$ constitutes a minimal pair of captions grounded in an image, the former being positive and the latter negative. Defining specificity with reference to both positive and negative details ensures that a model does not learn to simply assign higher similarity to longer captions, but evaluates the relevance of every new detail in the caption against the reference image.

\subsection{Specificity Rate}\label{specificity_rate}



To aggregate specificity across a set of minimal pairs, we introduce the \textbf{Specificity Rate} (SR). We define two variants: $\text{SR}_{\text{pos}}$ measures the proportion of cases in which adding an additional relevant detail (positive detail unit) increases the similarity score with the image, while $\text{SR}_{\text{neg}}$ measures the proportion of cases in which adding an irrelevant detail (negative detail unit) decreases the similarity.
Given a set of $N$ positive or negative triplets, we compute the SR as follows:
\begin{equation}
\text{SR}_{\text{pos}} = \frac{1}{N}\sum_{j}^{N}\mathbb{I}[\theta(i,c_j + d_+) > \theta(i,c_j)]
\end{equation}
\begin{equation}
\text{SR}_{\text{neg}} = \frac{1}{N}\sum_{j}^{N}\mathbb{I}[\theta(i,c_j) < \theta(i,c_j  + d_-)]
\end{equation}

\noindent where $\mathbb{I}[\cdot]$ is the indicator function which outputs 1 if the condition inside is true and 0 otherwise, and $\theta$ is the cosine similarity between the representations of image and text. This formulation captures the rate at which representational similarity increases with added positive details, or decreases with added negative ones, thus measuring model specificity.


\subsection{Specificity-Aware Learning}

Although specificity can be used purely for evaluation, we can also enforce it during training. To encourage the model to prefer captions that describe images with greater relevant detail, we introduce a training objective that rewards higher similarity scores for incrementally more informative captions, and lower similarity scores for less accurate ones. Given a dataset of $N$ positive and $N$ negative triplets, we define the following hinge loss with a dynamic margin:

{\small
\begin{equation}
\mathcal{L}_{\text{pos}} = \frac{1}{N} \sum_{i}^{N} \max\left(0, \theta(i,c_j) - \theta(i,c_j + d_+) + \epsilon \right),
\end{equation}
}

\noindent where $\epsilon$ is a batch-wise average similarity difference between positive and base captions, which is detached from gradient computation and clamped for numerical stability:

\[
\epsilon = \text{detach}\left(\frac{1}{N} \sum_{i}^{N} \left( \theta(i,c_j + d_+) - \theta(i,c_j)\right)\right),
\]

The negative loss, $\mathcal{L}_{\text{neg}}$ is computed by analogy, from the negative triplets in the dataset.



The final training objective combines the contrastive loss, the positive detail loss and the negative detail loss:

\begin{equation}
\label{eq:main_loss}
\mathcal{L} = \alpha \mathcal{L}_{\text{contrastive}} + \beta \mathcal{L}_{\text{pos}} + \gamma \mathcal{L}_{\text{neg}},
\end{equation}

\noindent where $\alpha$, $\beta$, and $\gamma$ are weighting hyperparameters tuned on a validation set. Figure~\ref{fig:model-fig} illustrates the overall training framework. 

\subsection{Metric Computation}

Given a SPecificity-Enhanced CLIP (SPEC) model trained as described above, we score candidate captions, $\hat{c}$, against input image, $i$, as follows:


\begin{equation}
\mathrm{SPECS} = \theta_{\mathrm{norm}}(i,\hat{c})
\end{equation}

\

\noindent i.e., the metric uses cosine similarity, clipped at 0.

\section{Experiments and Evaluation}

\subsection{Training and Validation Datasets}
We train our model on the ShareGPT-4V dataset~\cite{chen2024sharegpt4v}, which contains 1.2 million high-quality image-caption pairs synthetically generated by a strong captioning model, instructed to mention object attributes, spatial layouts, and aesthetic properties. The images in the dataset are sourced from COCO~\cite{lin2014microsoft}, SAM~\cite{kirillov2023segment}, and LAION~\cite{schuhmann2022laion}, and captions are 143 tokens long on average. 

For intrinsic specificity evaluation, we use the sDCI dataset~\cite{urbanek2024picture}, consisting of 7805 images, each paired with 10 captions, which are synthetically designed to fit in CLIP's context window of 77 tokens. This underutilizes the full context window of our model, but enables controlled comparisons to other models, constrained by the 77-token context window. We compare against the models introduced in Section~\ref{sec:rw}, as well as SigLIP~\cite{zhai2023sigmoid}, a recent vision-language model, which replaces the softmax contrastive loss in CLIP with a pairwise sigmoid loss, improving efficiency, scalability and zero-shot transfer.

\subsection{Data Preprocessing}

To create the data needed to measure and train for specificity, we build a pipeline that segments captions into detail units.

\paragraph{Main Logic}
We considered various methods to segment captions into detail units, based on part-of-speech tagging, dependency parsing and coreference resolution: the results were either unsatisfactory, slow to obtain or obstructed by technical challenges with the deployment of outdated libraries. The solution that proved best in terms of speed, ease of implementation and quality was obtained with the help of GPT-4. We presented the model with an example of a manually annotated caption and had it generate Python code that implements the segmentation pattern found in the example. The resulting code is based on part-of-speech tagging and a rule-based grammar (see Appendix~\ref{sec:appendix_grammar}).
Through manual inspection, we established that the solution is largely effective, but it somewhat oversegments the captions.

\paragraph{False Negatives vs. False Positives}
Considering the intended use of the segmented data, we determined that allowing for false negatives (i.e., missing splits) is less harmful than introducing false positives (i.e., incorrect extra splits). In other words, we prefer case (a), where a possible split is missing, over case (b), where an erroneous split is inserted:

\begin{itemize}
\item[(a)] \textit{A front view of a statue on cement} \textbar\ \textit{in a park.}
\item[(b)] \textit{A front} \textcolor{red}{\textbar}\ \textit{view of a statue} \textbar\ \textit{on cement} \textbar\ \textit{in a park.} 
\end{itemize}

Our goal is to ensure that every detail unit contains meaningful and novel information, and preserves the grammaticality of the caption. False positives introduce noise that may corrupt the metric signal and compromise training, especially when such errors accumulate.

Given the above reasoning, we modify the segmentation code with several rules to avoid splitting off (1) sentence-initial noun phrases that begin with \textit{The} as they are likely to repeat a previously mentioned entity, (2) prepositional phrases from the noun phrase preceding them as they are likely a modifier to the noun phrase, often referring back to previously mentioned objects (e.g. \textit{The cat} \textit{is partially ...} in Figure~\ref{fig:first_intuition}), (3) segments which start with a prepositional phrase from the context that follows, unless the segment contains a verb, as they are likely a location modifier to the following noun phrase (e.g. \textit{To the left of the car} \ \textit{there is a box}).\footnote{Sometimes, this rule would result in a false negative.}

\paragraph{Negative Triplets} For every positive triplet $\{i, c_j, c_j + d_+\}$, we create a negative counterpart, $\{i, c_j, c_j + d_-\}$, by randomly sampling a detail unit from another image-caption pair in the batch. This technique results in what could be called \textit{easy} negatives, i.e. random negatives which can be easily identified as irrelevant to the reference image. Future work could explore the use of hard negatives, but in this work we find that with the right loss weight balancing (see Section~\ref{sec:ablations}), even this weaker signal can be leveraged effectively. 

\subsection{Experimental Setup}

We train a base LongCLIP-B/32 model with a context window of 248, using standard contrastive training for six epochs. The best checkpoint is then fine-tuned with our specificity objective (see Eq.~\ref{eq:main_loss}), for another three epochs.

We use the Adam optimizer with a learning rate of $1\times10^{-5}$, weight decay of $1\times10^{-2}$, a batch size of 100 per GPU, and gradient accumulation over 4 steps (yielding an effective batch size of 400 per GPU). We set the loss weights to $\alpha=1$, $\beta=8$, and $\gamma=0.8$ based on extensive hyperparameter tuning. 
All experiments are conducted on four NVIDIA A40 GPUs. Training the model requires approximately one hour per epoch (4 GPU hours).

\begin{table}[t]
  \centering
  \resizebox{\linewidth}{!}{
  \begin{tabular}{l|ccc}
    \toprule
    \textbf{Model} & \textbf{Positive} & \textbf{Negative} & \textbf{Average} \\
    \midrule
    CLIP & 62.61 & 68.28 & 65.44 \\
    LongCLIP & 60.12 & 69.93 & 65.02 \\
    SigLIP & 58.56 & 76.28 & 67.42 \\
    NegCLIP & 54.96 & 78.84 & 66.90 \\
    DCI & 55.68 & 63.63 & 59.66 \\
    DAC & 46.84 & 66.88 & 56.86 \\		
    LaCLIP & 60.98 & 68.82 & 64.90 \\
    TripletCLIP & 53.34 & 70.20 & 61.77 \\
    \midrule
    LongCLIP* & 58.64 & 77.03 & 67.83 \\
    \textbf{SPEC} & \textbf{95.37} & \textbf{90.37} & \textbf{92.87} \\
    \bottomrule
  \end{tabular}
  }
  \caption{
    Specificity performance of various vision-language models on the sDCI dataset. 
    Positive and Negative correspond to $\text{SR}_{\text{pos}}$ and $\text{SR}_{\text{neg}}$ as defined in Section~\ref{specificity_rate}.
    \textit{LongCLIP} refers to the ViT-B/16 model as reported in the original paper, while \textit{LongCLIP*} is a model we trained from ViT-B/32. 
  }
  \label{tab:sDCI}
\end{table}

\begin{table*}[t] 
  \centering
  \resizebox{.98\textwidth}{!}{
  \begin{tabular}{l|c|c|c|c|c|c|c}
    \toprule
    \textbf{Metric} & \textbf{PCC $\rho$~$\uparrow$} & \textbf{1 - R$^2$~$\downarrow$} & \textbf{Kd $\tau$~$\uparrow$} & \textbf{Sp $\tau$~$\uparrow$} & \textbf{Base Model} & \textbf{Reference Free} & \textbf{TFLOPs} \\
    \midrule
    \multicolumn{8}{c}{\textbf{Rule-Based Evaluation}} \\
    \midrule
    BLEU-4     & \textcolor{gray}{0.3439} & \textcolor{gray}{62.78}     & \textcolor{gray}{0.2693} & \textcolor{gray}{0.2931} & - & \xmark & - \\
    ROUGE       & \textcolor{gray}{0.2509} & \textcolor{gray}{156.05}    & \textcolor{gray}{0.1886} & \textcolor{gray}{0.1893} & - & \xmark & - \\
    METEOR     & \textcolor{gray}{0.3593} & \textcolor{gray}{111.95}    & \textcolor{gray}{0.2417} & \textcolor{gray}{0.2536} & - & \xmark & - \\
    CIDEr      & \textcolor{gray}{0.0522} & \textcolor{gray}{3.30E+07}  & \textcolor{gray}{0.0635} & \textcolor{gray}{0.0601} & - & \xmark & - \\
    \midrule
    \multicolumn{8}{c}{\textbf{Representational Similarity Evaluation}} \\
    \midrule
    SPICE           & \textcolor{gray}{0.2218} & \textcolor{gray}{156.11} & \textcolor{gray}{0.1731} & \textcolor{gray}{0.1907} & - & \cmark & - \\
    CLIPScore      & \textcolor{gray}{0.2183} & \textcolor{gray}{26.04}  & \textcolor{gray}{0.1724} & \textcolor{gray}{0.1480} & CLIP & \cmark & $1.48 \times 10^{-2}$ \\
    PACScore       & \textcolor{gray}{0.1525} & \textcolor{gray}{20.93}  & \textcolor{gray}{0.1117} & \textcolor{gray}{0.1260} & CLIP & \cmark & $1.48 \times 10^{-2}$ \\
    LaCLIP         & \textcolor{gray}{0.1177} & \textcolor{gray}{71.94}  & \textcolor{gray}{0.0911} & \textcolor{gray}{0.1192} & CLIP & \cmark & $1.48 \times 10^{-2}$ \\
    TripletCLIP     & \textcolor{gray}{0.1697} & \textcolor{gray}{34.70}  & \textcolor{gray}{0.0852} & \textcolor{gray}{0.1038} & CLIP & \cmark & $1.48 \times 10^{-2}$ \\
    NegCLIP         & \textcolor{gray}{0.0872} & \textcolor{gray}{131.57} & \textcolor{gray}{0.0623} & \textcolor{gray}{0.0256} & CLIP & \cmark & $1.48 \times 10^{-2}$ \\
    LongCLIP       & \textcolor{gray}{0.2320} & \textcolor{gray}{18.58}  & \textcolor{gray}{0.1769} & \textcolor{gray}{0.2603} & LongCLIP & \cmark & $2.81 \times 10^{-2}$ \\
    LongCLIP*      & \textcolor{gray}{0.1723} & \textcolor{gray}{33.67}  & \textcolor{gray}{0.1484} & \textcolor{gray}{0.1662} & LongCLIP & \cmark & $2.81 \times 10^{-2}$ \\
    \rowcolor{blue!20}SPECS (Ours) & 0.5228 & 3.65 & 0.4078 & 0.5400 & LongCLIP & \cmark & $2.81 \times 10^{-2}$ \\
    \midrule
    \multicolumn{8}{c}{\textbf{LLM-Based Evaluation}} \\
    \midrule
    FaithScore      & \textcolor{gray}{0.1937} & 3.22  & \textcolor{gray}{0.1626} & \textcolor{gray}{0.1115} & LLaMA & \cmark & 3.97 \\
    CLAIR           & \textcolor{gray}{0.3815} & 1.98  & \textcolor{gray}{0.3847} & \textcolor{gray}{0.4552} & Claude & \cmark & - \\
    GPT4-Eval       & \textcolor{gray}{0.3976} & 2.95  & \textcolor{gray}{0.3447} & \textcolor{gray}{0.3866} & GPT-4 & \cmark & - \\
    RLAIF-V         & \textcolor{gray}{0.3547} & \textcolor{gray}{5.32}  & \textcolor{gray}{0.2774} & \textcolor{gray}{0.2544} & LLaVA & \cmark & 3.97 \\
    CAPTURE         & \textcolor{gray}{0.3521} & \textcolor{gray}{7.62}  & \textcolor{gray}{0.2801} & \textcolor{gray}{0.3449} & InternVL & \xmark & 4.54 \\
    FLEUR           & \textcolor{gray}{0.4230} & 3.01  & 0.4246 & \textcolor{gray}{0.5325} & LLaVA & \cmark & 7.37 \\
    DCSCORE         & 0.6605 & 1.54 & 0.5328 & 0.6166 & GPT-4o & \xmark & - \\
    \bottomrule
  \end{tabular}
}
\caption{Correlation of image captioning evaluation metrics and human judgments: \textnormal{Pearson’s $\rho$ (PCC $\rho$), $1 - R^2$, Kendall’s $\tau$ (Kd $\tau$), and Spearman’s $\tau$ (Sp $\tau$).For a fair comparison of computational cost, LLM-Based evaluations were computed using an input sequence length of 300 tokens, matching the setting used for Model-Based metrics. Correlation scores lower than those for SPECS are displayed in gray.
All $p$-values are less than 0.001.}}
  \label{human-correlation}
\end{table*}


\subsection{Results} 

\paragraph{Intrinsic Evaluation}

To evaluate whether our training objective effectively enhances specificity, we compare the specificity rate of SPEC (the base model behind the SPECS metric) against various vision-language models (see Section~\ref{specificity_rate}) on the sDCI benchmark and report results in Table~\ref{tab:sDCI}. SPEC achieves the best performance across all VLM models, with $\text{SR}_{\text{pos}} = 95.37$ and $\text{SR}_{\text{neg}} = 90.37$, resulting in an average specificity score of 92.87. Compared to the LongCLIP* baseline of 67.83, our model yields a substantial improvement of +25.04 points. The largest gain appears in $\text{SR}_{\text{pos}}$, where SPEC outperforms LongCLIP* by +36.73, highlighting its superior ability to recognize and prefer more detailed captions, and the effectiveness of the custom training objective. 


Interestingly, models with strong general-purpose performance do not necessarily achieve high specificity scores. For example, SigLIP, despite strong results on standard vision-language benchmarks, performs worse than CLIP-based variants in both $\text{SR}_{\text{pos}}$ and average specificity. This suggests that model architecture alone is not sufficient to capture fine-grained image-text alignment. Models designed to improve compositionality show mixed results: NegCLIP slightly improves over CLIP, while DCI and DAC perform worse, and LaCLIP shows no improvement over the baseline.

Having established that the model we trained shows strong specificity in intrinsic evaluations, we proceed to use this model as a scoring function.

\paragraph{Extrinsic Evaluation}

To evaluate how well automatic caption metrics align with human preferences, we adopt the evaluation protocol from DECAPBENCH~\cite{ye2025painting}. This human correlation benchmark consists of 100 images, sampled from the ImageInWords (IIW) dataset~\cite{garg2024imageinwords}. Human-annotated ratings are available for five captions per image, generated by different vision-language models. This setup enables a standardized comparison between automatic metrics and human judgments. We evaluate SPECS in the context of a wide range of metrics from different categories: rule-based, representational similarity-based and LLM-based ones. 
Table~\ref{human-correlation} summarizes the results in terms of four standard correlation metrics: Pearson correlation coefficient (PCC), coefficient of determination ($R^2$), Kendall’s $\tau$ (Kd $\tau$), and Sample-wise $\tau$ (Sp $\tau$). 

Among RS metrics, SPECS achieves the highest human correlation, improving PCC over the CLIPScore baseline from 0.2183 to 0.5228 and Kendall’s $\tau$ from 0.1724 to 0.4078 . SPECS outperforms most LLM-based metrics, including the strongest open-source metric, FLEUR, in terms of PCC and ranking consistency (Sp $\tau$).

SPECS requires only $2.81 \times 10^{-2}$ TFLOPs per forward pass, making it far more efficient than LLM-based metrics like FLEUR (7.74) and CLAIR (3.97). With only 0.15 billion parameters, SPECS remains lightweight and scalable, offering a practical and human-aligned solution for evaluating dense, detail-rich captions.

\begin{table*}[t]
  \centering
  \resizebox{\textwidth}{!}{
  \begin{tabular}{l|cc|cc|ccc}
    \toprule
    \textbf{Model} & \multicolumn{2}{c|}{\textbf{Urban-1k}} & \multicolumn{2}{c|}{\textbf{COCO}} & \multicolumn{3}{c}{\textbf{Classification}} \\
     & Text-Image & Image-Text & Text-Image & Image-Text & ImageNet & CIFAR-10 & CIFAR-100 \\
    \midrule
    CLIP & 47.10 & 61.10 & 30.45 & 50.40 & 68.40 & 89.75 & 64.20 \\
    LongCLIP & 79.30 & 79.20 & 40.40 & 57.63 & 66.80 & 90.69 & 69.30 \\
    
    SigLip & 62.40 & 63.10 & 47.18 & 65.34 & 76.08 & 92.44 & 72.59 \\
    NegCLIP & 52.80 & 55.60 & 41.56 & 56.86 & 55.84 & 85.90 & 60.90 \\
    DCI & 43.00 & 29.70 & 21.44 & 20.55 & 53.34 & 87.38 & 57.96 \\
    DAC & 23.60 & 11.40 & 37.53 & 33.49 & 52.36 & 89.86 & 64.04 \\
    \midrule
    LongCLIP* & 77.00 & 75.80 & 35.50 & 52.44 & 59.91 & 90.38 & 66.36 \\
    SPEC & 69.80 & 0.30 & 22.72 & 4.48 & 11.01 & 71.26 & 33.10 \\
    \bottomrule
  \end{tabular}
  }
  \caption{
    Performance of different VL representation models on standard retrieval and classification benchmarks. Results cover long- and short-caption text–image retrieval (Urban-1k, COCO) and image classification (ImageNet, CIFAR-10, CIFAR-100). This table illustrates how specificity-oriented training, while improving fine-grained alignment, can impact general-purpose performance. 
    }
  \label{tab:vlm_eval}
\end{table*}

\section{Further Analysis}

\subsection{Hubness in the Embedding Space}

Although our specificity-enhanced model excels at fine-grained alignment, we observe a decline in performance on standard vision-language tasks such as retrieval and classification. We evaluate generalization across a diverse set of benchmarks, including Urban-1k~\cite{zhang2024long} and COCO~\cite{lin2014microsoft} for text-image retrieval, and ImageNet~\cite{russakovsky2015imagenet}, CIFAR-10, and CIFAR-100~\cite{krizhevsky2009cifar} for image classification. These benchmarks span both multimodal and unimodal settings, providing a comprehensive view of how specificity-oriented training impacts general-purpose representations (Table~\ref{tab:vlm_eval}).

Specifically, our training objective modifies the geometry of the embedding space by introducing additional constraints beyond contrastive similarity, particularly encouraging alignment with incrementally detailed captions. While this improves the model’s specificity, it disrupts the isotropy of the representation space and results in the emergence of hubness: caption embeddings that are overly similar to many images, ultimately degrading retrieval performance.

Overall, while our training strategy enhances specificity evaluation, it can distort the geometry of the embedding space, negatively affecting performance on other downstream tasks. This does not devalue the SPECS metric, but sheds some light into the mechanism it adopts to provide reliable evaluation scores for long image captions. 

\subsection{Compositionality Analysis}


While our main focus is on improving specificity, we also explore whether it leads to improve compositional reasoning. It is reasonable to expect that models capable of handling variations in attribute order or relational structure may also perform better on incrementally positive descriptions. To test this, we evaluate our models on two established benchmarks: ARO~\cite{yuksekgonul2022and}, which measures understanding of attribute-relation-object structure and word order sensitivity, and SugarCREPE++ (SCPP)~\cite{dumpala2024sugarcrepe++}, which assesses sensitivity to semantic equivalence under lexical variation. Results are shown in Table~\ref{tab:aro-scpp}, with full SCPP details in Appendix~\ref{sec:scpp}.

On ARO, SPEC exhibits considerably higher performance than LongCLIP*, which suggests a direct relationship between specificity and compositionality. This finding does not hold on the SCPP benchmark, however. In fact among all compositionality-enhanced models, only NegCLIP shows a marked improvement on SCPP over the base CLIP model, all others either matching the base performance or showing a considerable degradation (e.g., DAC.)

\begin{table}[t]
  \centering
  \resizebox{\linewidth}{!}{
  \begin{tabular}{l|cccc|c}
    \toprule
    \textbf{Model} & \textbf{Rel.} & \textbf{Attr.} & \textbf{C/O} & \textbf{F/O} & \textbf{SCPP} \\
    \midrule
    CLIP & 59.84 & 63.96 & 47.28 & 58.54 & 53.33 \\
    LongCLIP & 59.70 & 63.42 & 56.91 & 69.03 & 54.45 \\
    SigLip & 46.52 & 56.24 & 32.95 & 40.86 & 20.88 \\
    NegCLIP & 70.52 & \textbf{81.08} & 87.04 & 90.38 & \textbf{63.79} \\
    DCI & 81.31 & 73.85 & \textbf{94.53} & \textbf{95.68} & 51.29 \\
    DAC & \textbf{76.18} & 67.63 & 88.58 & 91.25 & 43.54 \\
    La-CLIP & 45.48 & 58.72 & 34.97 & 40.54 & 54.99 \\
    TripletCLIP & 54.94 & 63.07 & 23.53 & 27.58 & 55.71 \\
    \midrule
    LongCLIP* & 52.96 & 65.81 & 63.97 & 70.20 & 56.74 \\
    SPEC & 73.38 & 69.31 & 75.23 & 84.96 & 35.61 \\
    \bottomrule
  \end{tabular}
  }
  \caption{Performance of various models on the ARO and SCPP benchmarks.
  \textnormal{C/O and F/O correspond to compositionality evaluation on COCO-Order and Flickr30k-Order, respectively. }}
  \label{tab:aro-scpp}
\end{table}

\subsection{Caption Length Sensitivity}

We further examined how evaluation performance varies across different caption lengths. To this end, we divided captions into buckets based on token counts and measured human correlation within each range. As shown in Figure~\ref{fig:combined}, CLIPScore performs reasonably well on short captions (fewer than 60 tokens), while SPECS  yields consistently stronger correlation for longer captions. This trend suggests that short and long captioning represent two distinct regimes that require different evaluation focuses. Despite attempts to train a unified model that performs well across all lengths, we find a clear trade-off that limits joint performance. We therefore recommend using CLIP for captions under 60 tokens and SPEC for longer ones. The detailed results are provided in Appendix~\ref{sec:appendix_correlation}.

\begin{figure}[t]
    \centering
    \includegraphics[width=1\textwidth]{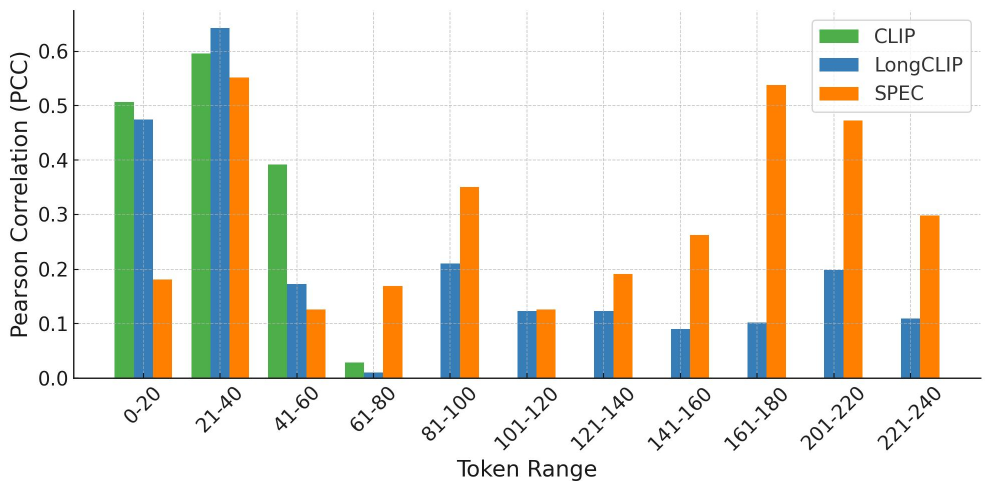}
    \caption{Comparison of Pearson Correlation (PCC) across token ranges for different models on the Flickr30k and ImageInWords datasets.}
    \label{fig:combined}
\end{figure}

\subsection{Hyperparameters}
\label{sec:ablations}


We tuned four key hyperparameters: loss weight  $(\alpha, \beta, \gamma)$, learning rate, loss type, and dataset shuffle ratio. Table~\ref{ablation} summarizes the results.

The optimal setting $(\alpha=1, \beta=8, \gamma=0.8)$ achieves the highest specificity score of 92.87. Alternative configurations such as $1:9:0.8$ and $1:8:0.6$ result in noticeably lower performance, underscoring the model’s sensitivity to the precise relative weighting of different training objectives.

Interestingly, the optimal setting is highly imbalanced, placing much greater emphasis on the positive loss compared to the contrastive and negative loss components. We believe that this imbalance arises from the nature of our specificity-focused training setup: since the contrastive loss is already well optimized from the pretrained CLIP checkpoint, and the negative detail examples are relatively easy, the model benefits more from strong and consistent supervision on the positive detail signal. The positive loss directly encourages the model to increase similarity for incremental, visually grounded additions---precisely the type of fine-grained distinction that we aim to capture. Thus, assigning a large weight to this component reinforces the core objective of our method.

\begin{table}[t]
  \centering
  \resizebox{\linewidth}{!}{
  \begin{tabular}{c|l|ccc}
    \toprule
    \textbf{Ablation} & \textbf{Config} & \textbf{Pos.} & \textbf{Neg.} & \textbf{Avg.} \\
    \midrule
    \multirow{3}{*}{\makecell{Loss Weight\\1:8:0.8}} 
    & 1:8:0.6 & 85.85 & 82.33 & 84.09 \\
    & 1:9:0.8 & 87.39 & 86.59 & 86.99 \\
    & 1:8:0.8 & 95.37 & 90.37 & 92.87 \\
    \midrule
    \multirow{3}{*}{\makecell{Learning Rate\\$1 \times 10^{-5}$}} 
    & $1 \times 10^{-6}$  & 77.22 & 68.64 & 72.93 \\
    & $5 \times 10^{-6}$  & 83.57 & 76.54 & 80.55 \\
    & $1 \times 10^{-5}$  & 95.37 & 90.37 & 92.87 \\
    \midrule
    \multirow{2}{*}{\makecell{Loss Type \\ hinge}} 
    & esp$1 \times 10^{-3}$   & 80.18 & 65.80 & 72.99 \\
    & hinge     & 95.37 & 90.37 & 92.87 \\
    \midrule
    \multirow{3}{*}{\makecell{ Shuffle Rate \\ 90\%}} 
    & 50\%  & 83.12 & 87.33 & 85.22 \\
    & 100\% & 87.27 & 83.11 & 85.19 \\
    & 90\%  & 95.37 & 90.37 & 92.87 \\
    \bottomrule
  \end{tabular}}
  \caption{Hyperparameter tuning.  
 }
  \label{ablation}
\end{table}


We also investigate the role of shuffle ratios when constructing negative captions. Since each negative caption is created by appending a detail unit sourced from other images in the batch, to the current base caption, using unshuffled units may result in semantically coherent and fluent text that unintentionally resembles a valid caption. This risks introducing false negatives that confuse the model during training.
To address this, we introduce a shuffle ratio hyperparameter that controls the proportion of negative detail units that are randomly shuffled at the token level before being appended. We find that a shuffle of 90\% yields the best performance. A ratio of 90\% means most units are shuffled to break semantic coherence, while a small portion (10\%) remain in their original order to preserve some challenging cases. 
This high optimal shuffle rate suggests that introducing controlled noise into the negatives improves the model’s ability to focus on genuine detail alignment without being misled by surface-level fluency.


\section{Conclusion}
The evaluation of long, detailed captions is a challenge with a pronounced quality-to-cost trade-off. We introduce specificity, a critical dimension for evaluating added detail in image captions. By fine-tuning a CLIP model with a specificity-aware learning objective, we develop SPECS, a new evaluation metric based on representational similarity. Extensive experiments demonstrate that SPECS strongly correlates with human judgments while remaining computationally efficient and scalable. 



\section*{Limitations}

While SPECS offers strong alignment with human judgments and excels at evaluating fine-grained visual grounding, its performance on standard vision-language tasks is limited. As shown in compositionality benchmarks such as ARO and SCPP, improvements in specificity do not directly translate into better reasoning over attribute structures or lexical variations. This indicates that the specificity-focused objective does not generalize well to tasks requiring structural or semantic invariance.

In addition, our hubness analysis reveals distortions in the embedding space caused by specificity-aware training. By encouraging sensitivity to visual details, the model tends to over-align with frequent or stylistically similar captions, leading to degraded performance in retrieval and classification tasks. These findings highlight a trade-off between detail sensitivity and general purpose utility.

Addressing this trade-off remains an open challenge. Future work may consider architectural modifications or auxiliary learning objectives that preserve fine-grained grounding while improving transferability to downstream tasks.


\bibliography{anthology}

\newpage

\appendix

\section{Segmentation Grammar}
\label{sec:appendix_grammar}

The following context-free grammar was used to define syntactic structures relevant to caption segmentation:

\begin{lstlisting}
grammar = r"""
    NP: {<DT>?<JJ.*>*<NN.*>+}              # Noun phrase with optional determiners and adjectives
    VP: {<VB.*><NP|PP|CLAUSE>+$}           # Verb phrase with verb followed by noun phrases, prepositional phrases, or clauses
    PP: {<IN><NP>}                         # Prepositional phrase with preposition followed by noun phrase
    CLAUSE: {<NP><VP>}                     # Clause containing noun phrase followed by verb phrase
    CONJ: {<CC><NP|VP|PP|CLAUSE>}          # Conjunction with conjoined structures
"""
\end{lstlisting}

\section{SCPP++ Result}\label{sec:scpp}

Table~\ref{tab:compositional_eval} presents the full results on the SCPP++ benchmark, broken down across five compositional variation types: Swap Object, Swap Attribute, Replace Relation, Replace Object, and Replace Attribute. Each variation is evaluated under two settings: ITT (Image-to-Text retrieval) and TOT (Text-Only Transfer), reflecting different forms of generalization stress.

Overall, we observe that models like NegCLIP and TripletCLIP maintain relatively strong performance across both ITT and TOT settings, while our SPEC model, although competitive in overall specificity evaluation, exhibits lower compositional generalization performance. This is consistent with earlier analysis in Section~5, and supports the claim that specificity-oriented fine-tuning does not necessarily improve compositional reasoning.

\section{Short Caption Human Correlation}
\label{sec:appendix_correlation}

\begin{table}
  \centering
  \resizebox{.80\textwidth}{!}{
  \begin{tabular}{l|c|c|c|c}
    \toprule
    \textbf{Metric} & \textbf{PCC $\rho$~$\uparrow$} & \textbf{1 - R$^2$~$\downarrow$} & \textbf{Kd $\tau$~$\uparrow$} & \textbf{Sp $\tau$~$\uparrow$} \\
    \midrule 
    \multicolumn{5}{c}{\textbf{COCO Correctness}} \\
    \midrule
    CLIP     & 0.6384 & 2.45  & 0.4987 & 0.6226 \\
    SPECS    & 0.3838 & 6.78  & 0.2952 & 0.4092 \\
    \midrule
    \multicolumn{5}{c}{\textbf{COCO Throughness}} \\
    \midrule
    CLIP     & 0.5785 & 2.98  & 0.4458 & 0.5790 \\
    SPECS    & 0.3645 & 7.53  & 0.2784 & 0.3989 \\
    \midrule
    \multicolumn{5}{c}{\textbf{Flickr8k Correctness}} \\
    \midrule
    CLIP     & 0.5328 & 3.52  & 0.4102 & 0.5422 \\
    SPECS    & 0.1228 & 66.7  & 0.1139 & 0.1451 \\
    \midrule
    \multicolumn{5}{c}{\textbf{Flickr8k Throughness}} \\
    \midrule
    CLIP     & 0.5012 & 4.00  & 0.3790 & 0.5421 \\
    SPECS    & 0.1995 & 25.12 & 0.1550 & 0.2193 \\
    \midrule
    \multicolumn{5}{c}{\textbf{Flickr30k Correctness}} \\
    \midrule
    CLIP     & 0.6071 & 2.71  & 0.4553 & 0.6219 \\
    SPECS    & 0.2299 & 18.9  & 0.1709 & 0.3058 \\
    \midrule
    \multicolumn{5}{c}{\textbf{Flickr30k Throughness}} \\
    \midrule
    CLIP     & 0.5352 & 3.49  & 0.4026 & 0.5805 \\
    SPECS    & 0.2230 & 20.12 & 0.1641 & 0.2769 \\
    \bottomrule
  \end{tabular}
  }
  \caption{\textbf{Human correlation in various short capiton datasets(COCO, Flickr8k, Flickr30k).}}
  \label{tab:specs_correlation}
\end{table}

To explore whether a unified evaluation model could perform well across both short and long captions, we evaluated SPECS and CLIP on datasets dominated by shorter captions on short caption datasets. As shown in Table~\ref{tab:specs_correlation}, CLIP consistently achieves higher correlation with human judgments across all datasets and metrics. This suggests that while SPECS is optimized for longer, detail-rich captions, it underperforms in short-caption settings. Our results indicate a clear trade-off, and confirm that a single model cannot simultaneously achieve optimal performance across all caption lengths.

\section{Model Code Names}

We provide the exact model code names used in our experiments to ensure reproducibility:
\begin{itemize}
    \item \textbf{CLIP}: \texttt{openai/clip-vit-base-patch32}
    \item \textbf{LongCLIP}: \texttt{BeichenZhang/LongCLIP-B}
    \item \textbf{SPECS}: \texttt{Xiaohud/SPECS}
    \item \textbf{FaithScore}: \texttt{llama2/llama-2-7b-hf}
    \item \textbf{CLAIR}: Claude Instant
    \item \textbf{GPT-4 Eval}: \texttt{gpt-4-0613}
    \item \textbf{RLAIF-V}: \texttt{llava-hf/llava-v1.6-7b-hf}
    \item \textbf{CAPTURE}: \texttt{OpenGVLab/InternVL2\_5-8B}
    \item \textbf{FLEUR}: \texttt{llava-hf/llava-v1.6-13b-hf}
    \item \textbf{DCSCORE}: \texttt{gpt-4o-2024-08-06}
\end{itemize}

\begin{table*}[t]
  \centering
  \resizebox{\textwidth}{!}{
  \begin{tabular}{l|cc|cc|cc|cc|cc|c}
    \toprule
    \textbf{Model} 
    & \multicolumn{2}{c|}{Swap Object} 
    & \multicolumn{2}{c|}{Swap Attribute} 
    & \multicolumn{2}{c|}{Replace Relation} 
    & \multicolumn{2}{c|}{Replace Object} 
    & \multicolumn{2}{c|}{Replace Attribute} 
    & Avg. \\
    & ITT & TOT & ITT & TOT & ITT & TOT & ITT & TOT & ITT & TOT & \\
    \midrule
    CLIP            & 45.18 & 19.74 & 45.21 & 33.03 & 56.26 & 38.62 & 86.80 & 83.72 & 65.61 & 59.14 & 53.31 \\
    Long-CLIP       & 42.85 & 15.10 & 49.39 & 31.98 & 55.68 & 40.54 & 90.19 & 87.71 & 71.31 & 59.77 & 54.45 \\
    Long-CLIP*      & 46.53 & 28.97 & 46.99 & 42.64 & 52.20 & 39.68 & 88.31 & 91.82 & 66.37 & 63.95 & 56.74 \\
    SigLIP          & 36.32 &  5.71 & 30.63 &  9.00 & 27.24 & 12.66 & 35.16 & 12.71 & 30.71 &  8.62 & 20.88 \\
    NegCLIP         & 55.25 & 34.65 & 57.99 & 56.47 & 52.27 & 51.57 & 89.53 & 94.55 & 69.41 & 76.27 & 63.79 \\
    DCI             & 44.10 & 31.80 & 45.60 & 38.00 & 43.20 & 35.70 & 80.20 & 81.20 & 60.90 & 52.20 & 51.29 \\
    DAC             & 27.80 & 11.40 & 33.50 & 25.40 & 48.60 & 48.60 & 64.30 & 75.80 & 44.00 & 56.00 & 43.54 \\
    La-CLIP         & 41.22 & 21.22 & 48.95 & 36.04 & 51.07 & 42.03 & 86.44 & 88.50 & 68.78 & 65.61 & 54.99 \\
    TripletCLIP     & 38.37 & 18.78 & 44.44 & 38.14 & 58.68 & 48.08 & 85.05 & 89.04 & 65.61 & 70.94 & 55.71 \\
    \midrule
    SPEC           & 30.61 & 16.73 & 28.37 & 24.02 & 25.96 & 24.25 & 48.36 & 73.91 & 38.57 & 45.30 & 35.61 \\
    \bottomrule
  \end{tabular}
  }
  \caption{
    \textbf{Compositional Generalization Evaluation.} 
    ITT and TOT denote image-to-text task and text-only task accuracy, respectively.
  }
  \label{tab:compositional_eval}
\end{table*}

\end{document}